\documentclass{osa-article}
\journal{boe}
\articletype{Research Article}

\usepackage{dblfloatfix}
\usepackage{multirow}
\usepackage[normalem]{ulem}
\usepackage{mathtools}
\usepackage[titletoc]{appendix}
\usepackage{booktabs}
\usepackage{makecell}
\usepackage{forest}
\usepackage{balance}
\usepackage{dblfloatfix}
\usepackage{soul}
\usepackage{array, tabularx, boldline}
\usepackage{graphicx}
\usepackage{cellspace}
\DeclareMathOperator{\SSIM}{SSIM}

\definecolor{aliceblue}{rgb}{0.94, 0.97, 1.0}
\def\code#1{\texttt{#1}}

\begin{document}

\title{DeepLSR: a deep learning approach for laser speckle reduction}

\author{Taylor L. Bobrow, Faisal Mahmood, Miguel Inserni, and Nicholas J. Durr\authormark{*}}

\address{Department of Biomedical Engineering, Johns Hopkins University (JHU), Baltimore, MD 21218, USA}

\email{\authormark{*}ndurr@jhu.edu} 


\begin{abstract}
Speckle artifacts degrade image quality in virtually all modalities that utilize coherent energy, including optical coherence tomography, reflectance confocal microscopy, ultrasound, and widefield imaging with laser illumination. We present an adversarial deep learning framework for laser speckle reduction, called DeepLSR (\url{https://durr.jhu.edu/DeepLSR}), that transforms images from a source domain of coherent illumination to a target domain of speckle-free, incoherent illumination. We apply this method to widefield images of objects and tissues illuminated with a multi-wavelength laser, using light emitting diode-illuminated images as ground truth. In images of gastrointestinal tissues, DeepLSR reduces laser speckle noise by 6.4 dB, compared to a 2.9 dB reduction from optimized non-local means processing, a 3.0 dB reduction from BM3D, and a 3.7 dB reduction from an optical speckle reducer utilizing an oscillating diffuser. Further, DeepLSR can be combined with optical speckle reduction to reduce speckle noise by 9.4 dB. This dramatic reduction in speckle noise may enable the use of coherent light sources in applications that require small illumination sources and high-quality imaging, including medical endoscopy.
\end{abstract}

\section{Introduction}

Laser illumination offers many advantages over incoherent light for imaging, including high power densities, efficient light generation, narrow spectral bandwidths, robust stability, long lifetimes, and fast triggering capabilities. Unfortunately, coherent illumination also introduces speckle artifacts that are caused by constructive and destructive interference between emitted wavefronts \cite{Goodman:76}. The poor image quality resulting from speckle noise prohibits lasers from being used in many widefield imaging applications. For example, commercial endoscopes utilize arc lamps or light-emitting diodes (LEDs) as illumination sources, and consequently require large-diameter light guides to transmit sufficient illumination power. Speckle noise also corrupts image quality in optical coherence tomography (OCT)  \cite{Bashkansky:00}, reflectance confocal microscopy \cite{rcm}, and ultrasound imaging \cite{Abbott1979}. To mitigate laser speckle noise, several optical methods have been explored \cite{GOODMAN2007,Liba2017}. In general, optical approaches add cost and complexity, reduce power throughput, and place fundamental limitations on imaging speed.

While image processing methods for laser speckle reduction have been most prominently developed for OCT \cite{Adler2004,OCTSalinas,OCTJian,OCTWong}, conventional model-based algorithms may be applied to speckle reduction in widefield imaging. In general, these denoising approaches can be grouped into three categories: total variation-based, non-local methods, and sparse filtering. Rudin \textit{et al.} first introduced the concept of Total Variation (TV) denoising \cite{TV}, which has been shown to work extremely well on piecewise-like images, but often suffers from a characteristic "staircase effect" in non-constant regions of images. Abergel \textit{et al.} expanded TV to include iterative conditional expectations \cite{TVICE}, and more recent work incorporated \textit{a priori} models for noise in TV \cite{TVMAP}. Multiplicative Image Denoising by Augmented Lagrangian (MIDAL) \cite{MIDAL} utilizes a multiplicative noise model (with knowledge of the noise standard deviation) and optimization through a Lagrangian framework, and generally mitigates noise with only some loss in image texture. Buades \textit{et al.} were the first to propose a non-local algorithm for noise reduction, called Non-Local Means (NLM) \cite{Buades}. NLM relies on the assumption that for each noisy feature in an image, there exists similar non-local features that may be combined to separate noise from the common, underlying image features. NLM has been adapted for a probabilistic Poisson noise model \cite{POISNLM}, and also paired with Principal Component Analysis (PCA) \cite{NLPCA} as well as the Wiener filter \cite{PNL}. Sparse filtering methods seek to maximize the variety of image features in a learned dictionary that is later averaged to reduce image noise. Aharon \textit{et al.} introduced a joint K-mean clustering and Singular Value Decomposition (K-SVD) approach \cite{KSVD}, which was further optimized by Rubenstein \textit{et al} \cite{AKSVD}. Block-Matching 3D Filtering (BM3D) and its variants are the de-facto state-of-the-art for image denoising using sparse filtering \cite{BM3D,BM3DCNN,NOISEREV}. Similar to K-SVD, BM3D relies on the aggregation of noisy blocks with comparable features for collaborative filtering and weighted averaging to reduce noise. This method was later expanded to RGB images by way of a transformation to the YUV image space \cite{CBM3D}. An in-depth review of prior work in relevant image denoising techniques is provided by Meiniel \textit{et al} \cite{NOISEREV}.

In general, conventional image processing techniques are computationally complex, require parameter tuning, and degrade resolution or introduce artifacts as they reduce speckle. Machine learning approaches, on the other hand, serve as an alternative to image processing techniques, as they can generate complex transformation functions by training on datasets that contain example input and desired output images. Moreover, deep learning has emerged as a powerful technique to learn complex representations of imaging data using multi-layer neural networks. Here, we present a deep convolutional neural network for laser speckle reduction (\textquotesingle \emph{DeepLSR}\textquotesingle) on widefield images formed from multi-wavelength, red-green-blue laser illumination. We describe a method for effectively learning the distribution of speckle artifacts to target and reduce noise in images not previously seen by the network. This technique relies on a training set of coherent- and incoherent-illuminated image pairs of a variety of objects to learn a transformation from speckled to speckle-free images. Previous efforts in OCT have explored shallow neural networks for estimating filter parameters in a speckle reduction model \cite{Avanaki2013}, and deep networks for speckle reduction using a set of registered and averaged volumes of retinal tissue as ground truth \cite{Ma2018}. CNNs have previously been shown to be effective for recovering image information degraded by scattering media \cite{ImageThroughDiffusers,DeepSpeckleCorrelation}. In widefield imaging, deep learning networks have been applied for general image denoising \cite{Agostinelli2013}, but not specifically for speckle reduction. DeepLSR is novel in its use of a true incoherent source as a target ground truth, the use of a diverse set of objects for training a generalizable model, and in its application of deep learning to widefield laser-illumination images. We benchmark this approach against conventional speckle reduction methods on images of laser-illuminated objects previously unseen by the network. We further provide step-by-step instructions for adapting DeepLSR to new data sets contaminated with speckle noise (see Appendix).

Standard deep learning models employ handcrafted loss functions that utilize repeated pixel-wise comparisons between the model\textquotesingle s prediction and the ground truth for model refinement. However, such pixel-wise loss functions do not capture higher order statistics that exist in the training data, such as non-local dependencies \cite{Richard}. To address relationships beyond the second order and to capture spatial relationships amongst distant pixels, recent focus has shifted to generative models to improve translations between higher dimensional data. Such methods have been used for image-to-image translation tasks with applications that include artistic style transfer \cite{artistic}, super-resolution imaging \cite{superres}, and synthetic data refinement \cite{synth}.

\begin{figure}[t!]
\centering
\includegraphics[width=11cm]{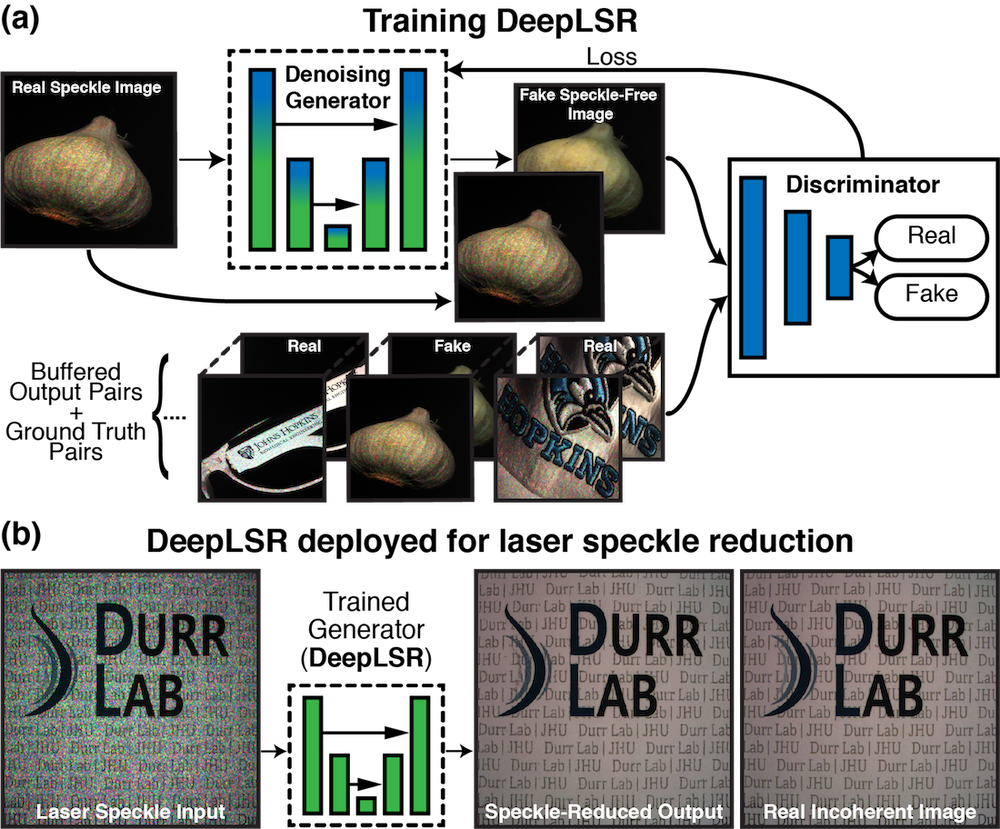}
\setlength{\belowcaptionskip}{-11pt}
\caption{DeepLSR Architecture. (a) Training architecture for image-to-image translation-based laser speckle reduction using a conditional Generative Adversarial Network. A generator learns to transform between pairs of images acquired with coherent and incoherent illumination while a discriminator learns to classify input images as real or fake. (b) Once training is complete, the discriminator is discarded and the trained generator (DeepLSR) reduces laser speckle noise in images not previously seen by the generator.}
\label{fig:Figure1}
\end{figure}

DeepLSR utilizes a conditional Generative Adversarial Network (cGAN) to reduce laser speckle by posing the problem as an image-to-image translation task \cite{Isola2016}. The overall architecture involves simultaneously training a speckle-free image generator and a real-versus-fake image discriminator, given a conditional input (Fig. \ref{fig:Figure1}). While the generator learns to generate a realistic mapping from an input speckled image to an output speckle-free image, the discriminator learns to classify pairs of input and generated output images as either real or fake. During this adversarial training, the discriminator provides feedback to the generator. The trained generator is then capable of reducing speckle noise in images it has never seen. Previous research in adversarial image-to-image translation has been challenged by instability caused by the complexity of back-and-forth training between the generator and the discriminator. Furthermore, this training paradigm can struggle with learning multiple sub-distributions of data that may exist within a certain distribution. This problem is often referred to as mode collapse. To overcome this challenge, we utilize spectral normalization instead of the more commonly used batch normalization for training DeepLSR \cite{Miyato2018}. With this adversarial framework, we trained networks to reduce speckle in images of a wide assortment of objects to evaluate robustness, and in images of tissue for evaluation in an endoscopic setting.

\begin{figure}[t!]
\centering
\includegraphics[width=12cm]{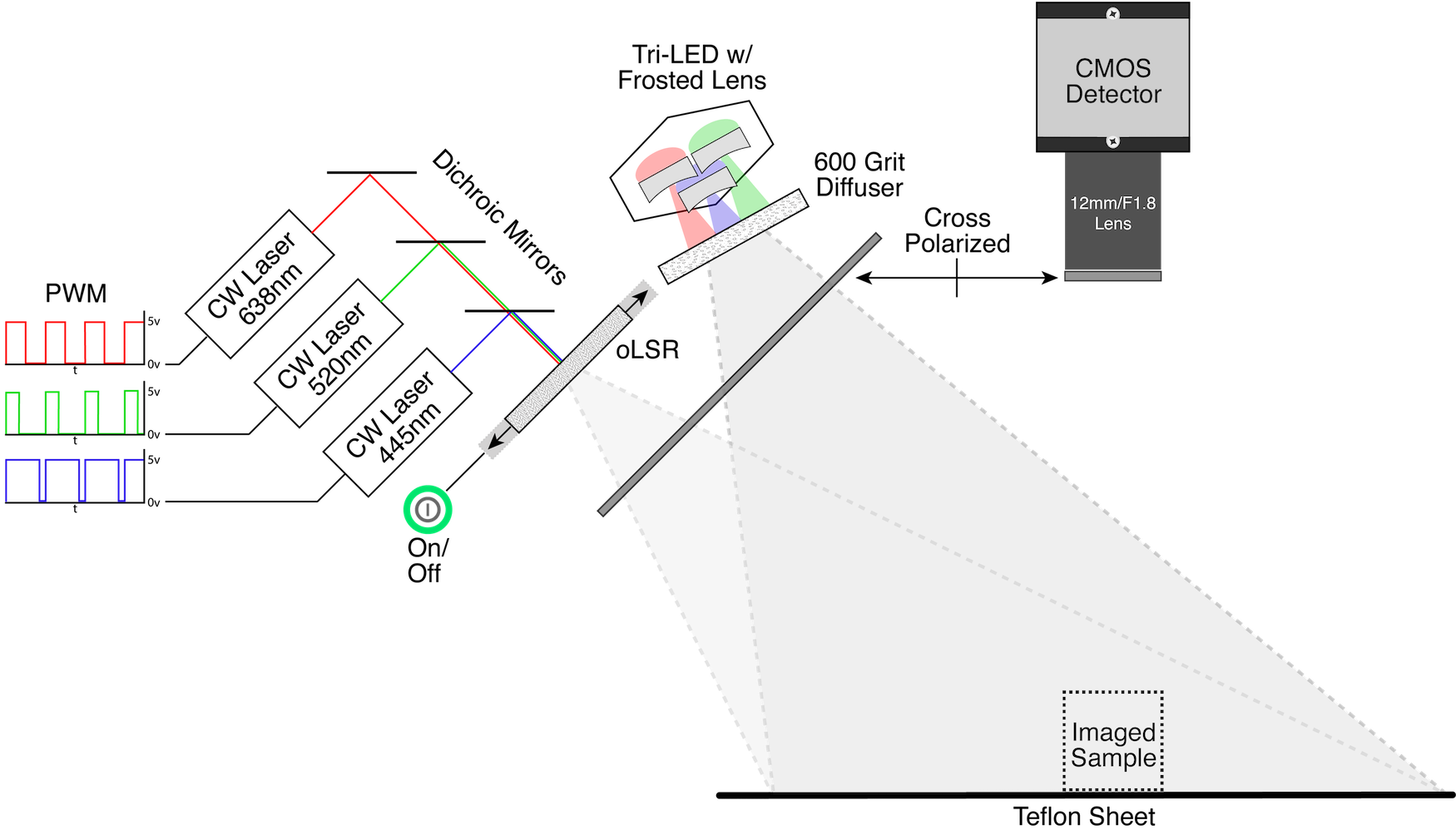}
\caption{Imaging setup for acquiring images with: (1) laser illumination (oLSR turned off), (2) laser illumination with optical laser speckle reduction (oLSR turned on), and (3) LED illumination.}
\label{fig:Figure2}
\end{figure}

\section{Methods}

We trained and tested DeepLSR using images acquired of assorted household and laboratory objects picked to represent a wide range of textures, shapes, and bidirectional reflectance distribution functions and images of \textit{ex-vivo} porcine esophagus, intestine, and stomach from three animals. These samples were illuminated using a red-green-blue laser for coherent illumination, the same laser with a commercial optical laser speckle reducer (oLSR) utilizing an oscillating diffuser, and an LED for incoherent illumination. We trained two networks to learn two transformations: (1) from laser illuminated images to LED illuminated images (DeepLSR), and (2) from optically laser speckle reduced images to LED illuminated images (DeepLSR+oLSR).

\subsection{Imaging Setup} Paired images for adversarial training were captured using laser illumination, laser illumination with optical speckle reduction, and light-emitting diode (LED) illumination \ref{fig:Figure2}. A laser unit with 445nm, 520nm, and 638nm diodes (Optlasers microRGB) and a white-light LED (Luxeon Star Tri-Star) with Rebel 448nm, 530nm, and 655nm diodes were used to illuminate samples. Diodes were selected so that wavelengths emitted by the laser and LED were similar. The laser unit\textquotesingle s beam was positioned normal to the aperture of an Optical Laser Speckle Reducer \cite{Graetzel2015} (\emph{oLSR}, Optotune LSR-3005-24D), and the oLSR was toggled on and off for imaging with and without speckle reduction. A frosted poly-carbonate triple lens (Luxeon Star \#10508) and a 600 grit diffusion lens were used to match the full-width at half maximum of the LED\textquotesingle s illumination intensity profile to that of the laser. For profile-matching, the laser and LED were both aimed at a Teflon imaging target placed below a color, 8-bit CMOS detector (ThorLabs \#DCC3240C, 12mm/F1.8) with an integration time set to 50 ms. A linear polarizing sheet (Thorlabs, \#LPVISE2X3) was placed in front of the light sources and a linear polarizer (Edmund Optics \#47316) was mounted to the detector and adjusted to minimize specular reflection by cross polarization. The illumination intensities of the laser and LED sources were matched by imaging the Teflon sheet and modulating the power of each individual laser diode color channel to achieve the same average pixel value as the corresponding LED channel. Periodic adjustments were made as the diodes\textquotesingle \ power drifted over time. The light source and detector positions were kept constant, and all image scenes were kept static while collecting each triad of images illuminated with laser light, laser light with optical speckle reduction, and LED illumination. Because the imaging setup was fixed when illumination was toggled, image registration and further calibration were not needed.

\subsection{Data Acquisition and Preprocessing} Data was acquired from: (1) 113 assorted household and laboratory objects imaged at up to 9 different positions, with each illumination source, resulting in 1533 images, and (2) 449 images of \textit{ex-vivo} porcine esophagus, intestine, and stomach from three animals, with each illumination source, resulting in 1347 images. The position of objects and tissues were varied using translation and rotation to collect multiple images of the same sample with different speckle interference patterns for both training and testing. Before training, the histograms of each laser and oLSR image were adjusted to match the corresponding LED image using uniform histogram matching to correct for any white-balancing discrepancies. The 2880 images were resized from 1280x1024 to 1024x1024 pixels using bicubic interpolation. Images were then divided into sets of laser, oLSR, and LED images to be paired with their corresponding ground truth images for network training. 90 images (39 images from 6 objects, and 51 images acquired from porcine tissue) were removed from the dataset for final network testing. The objects imaged for testing were not represented in the training set. Similarly, the porcine tissue images for testing came from a different animal than the images used for training.

\subsection{Network Training} An adversarial deep learning paradigm was used to train the networks. To this effect, two deep networks, a generator and a discriminator, were iteratively trained. The generator was tasked with generating target images and the discriminator was tasked with classifying the generator output as real or fake and giving feedback to the loss function of the generator. This paradigm enables the network to use non-local information when making determinations.

The network proposed here has two loss terms: a GAN loss which is updated every iteration based on the feedback from the discriminator and an $\ell_1$ loss. The $\ell_1$ loss compliments the GAN loss to minimize the difference between the output and ground truth. By inserting this term, the objective of the generator is not only to fool the discriminator but also to be close to the ground truth output. Recent work by Kuarch \textit{et al.} has demonstrated that in GANs, hand-crafting additional loss functions is not significantly beneficial to the overall adversarial training process \cite{LossCompare}. Because $\ell_2$ is more sensitive to outliers in the prediction, there is a possibility that it will try to adjust the model according to these outlier values, even at the expense of other well predicted samples. For this reason, we use an $\ell_1$ loss function, as do many GAN-based image-to-image translation methods \cite{Isola2016}. To prevent mode collapse, images were pooled and fed to the discriminator in batches rather than individual images in each iteration. Spectral normalization was used to stabilize GAN training when learning simultaneously from assorted objects and tissue. The problem was solved using \textit{Adam} for stochastic optimization \cite{Kingma2014}. Further details about the generator and discriminator architectures can be found in \cite{DBLP:journals/corr/RadfordMC15}.

The network was trained for 400 epochs. The learning rate was set to 0.0002 for the first 200 epochs and linearly decayed to a learning rate of zero over the remaining 200 epochs. The size of the image buffer that stores generated images was set to 64. The networks were implemented using PyTorch 0.4 and the training was run on Nvidia P100 GPUs using Google Cloud. The average training time for each epoch was 303 seconds and the entire network was trained in approximately 33.66 hours. Once the training process is complete, the trained network computes speckle-reduced images at 6 frames per second on a virtual workstation with 4 CPUs on a 2.6 GHz Intel Xeon E5 processor and at 27 frames per second when using a P100 GPU.

\subsection{Evaluation Metrics} To quantify the performance of DeepLSR, we measured the peak signal-to-noise (PSNR) ratio and Structural Similarity Index (SSIM) \cite{Wang2004} between computed images and the incoherent, speckle-free images. PSNR assesses the relative noise of an image and was computed using,

\begin{equation}
PSNR = 10 \log_{10} (\frac{R^2} {MSE})
\end{equation}

\noindent where the maximum possible intensity $R$ is 255 and $MSE$ is the mean squared error between images. SSIM is a valuable metric for image comparison using quantities that are important for human perception (image contrast, luminance, and structure). SSIM was computed using,

\begin{equation}
\SSIM(x,y) = \frac{(2\mu_x\mu_y + C_1) (2 \sigma _{xy} + C_2)} {(\mu_x^2 + \mu_y^2+C_1) (\sigma_x^2 + \sigma_y^2+C_2)}
\end{equation}

\noindent where $x$ and $y$ are images for comparison, the image mean is $\mu$, variance is $\sigma$, covariance between $x$ and $y$ is $\sigma _{xy}$, and the constant $C$ is added for avoiding instability when the denominator is close to 0. Average image SSIM was calculated using windows of 11x11 pixels. The resultant SSIM index is a value between -1 and 1, where an index of 1 indicates equivalent image inputs. Images of a slanted edge and of a 1951 United States Airforce Resolution Target were used to assess DeepLSR's effect on image resolution. The Slanted Edge MTF plugin available for ImageJ \cite{Schindelin2012} was used to compute modulation transfer functions.

\begin{figure}[!htbp]
\centering
\makebox[\textwidth][c]{\includegraphics[width=11.5cm]{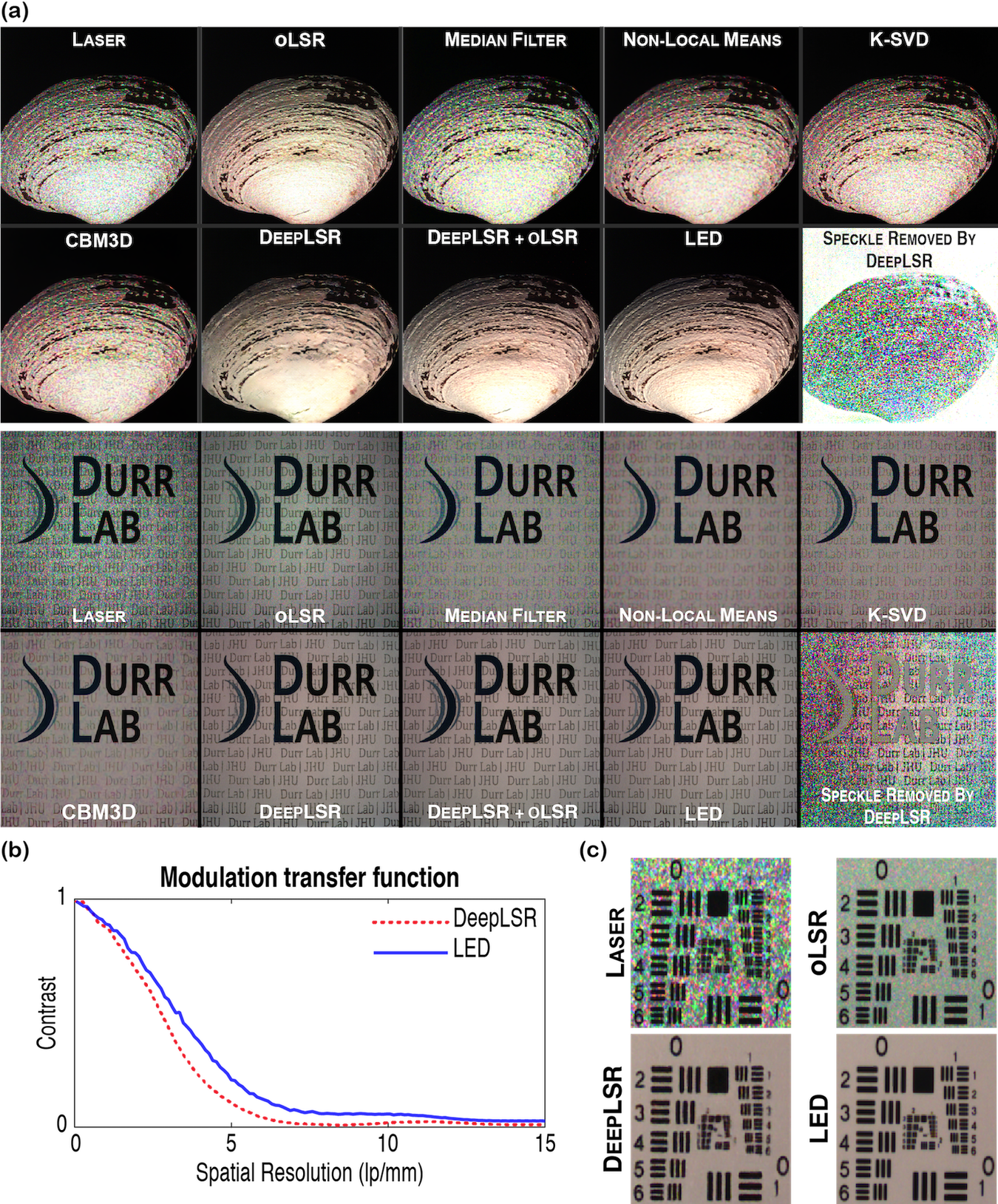}}
\caption{DeepLSR compared to conventional speckle reduction methods. DeepLSR was trained on an assortment of images that represent a variety of textures, shapes, and bidirectional reflectance distribution functions. (a) Images of two test objects illuminated with laser illumination, laser illumination with optical speckle reduction (oLSR), median filtering, non-local means, K-SVD, CBM3D, DeepLSR applied to the laser illuminated image, DeepLSR+oLSR applied to the optically speckle reduced image (DeepLSR+oLSR), the target speckle-free image illuminated with a light-emitting diode (LED), and the speckle artifacts removed from the laser illuminated image by DeepLSR. (b) Modulation transfer functions for LED illumination and laser illumination with DeepLSR found using a slanted edge. (c) Images of a 1951 United States Air Force Target with each illumination strategy and laser illumination with DeepLSR.}
\label{fig:Figure3}
\end{figure}

\begin{table}[b!]
	\begin{center}
	\begin{tabular}{p{3cm} p{9cm}} 
 	\hline
	\textbf{Method} & \textbf{Parameters} \\ [0.5ex]
 	\hline
		Median Filter 		& Kernel Size= 7 \\
		Non-Local Means     & {Kernel Size= 4, Window Size= 5, Filter Strength= 0.283} \\
		K-SVD		   	    & Block size= $[5x5x3]$, Dictionary Size= 1000, Training Blocks= 1000, $\sigma_{Noise}$= 0.01 \\
		CBM3D    	        & $\sigma_{Noise}$= 83.6 \\
		DeepLSR				& {Learning Rate= 0.002 (epochs 1-200) and linearly decayed (epochs 201-400), Pool Size= 64, patchGAN= $[70x70]$, $\lambda_{L1}$= 70} \\
	\hline
	\end{tabular}
	\caption{Parameters for noise reduction methods}
	\label{tab:Table1}
	\end{center}
\end{table}

\subsection{Performance} Both trained networks were evaluated using the reserved test images of assorted objects and porcine tissues. For performance benchmarks, we report PSNR and SSIM comparisons for laser vs. LED and oLSR vs. LED. We also compared DeepLSR to other common image processing denoising techniques: median filtering \cite{Huang1979}, non-local means \cite{Buades}, K-SVD \cite{KSVD}, and CBM3D \cite{CBM3D}. The input parameters for these algorithms were determined by multi-objective optimization, where PSNR and SSIM were the objective functions, using the same training data set as was used for DeepLSR (Table \ref{tab:Table1}).

\section{Results}

\begin{figure}[!t]
\makebox[\textwidth][c]{\includegraphics[width=11.5cm]{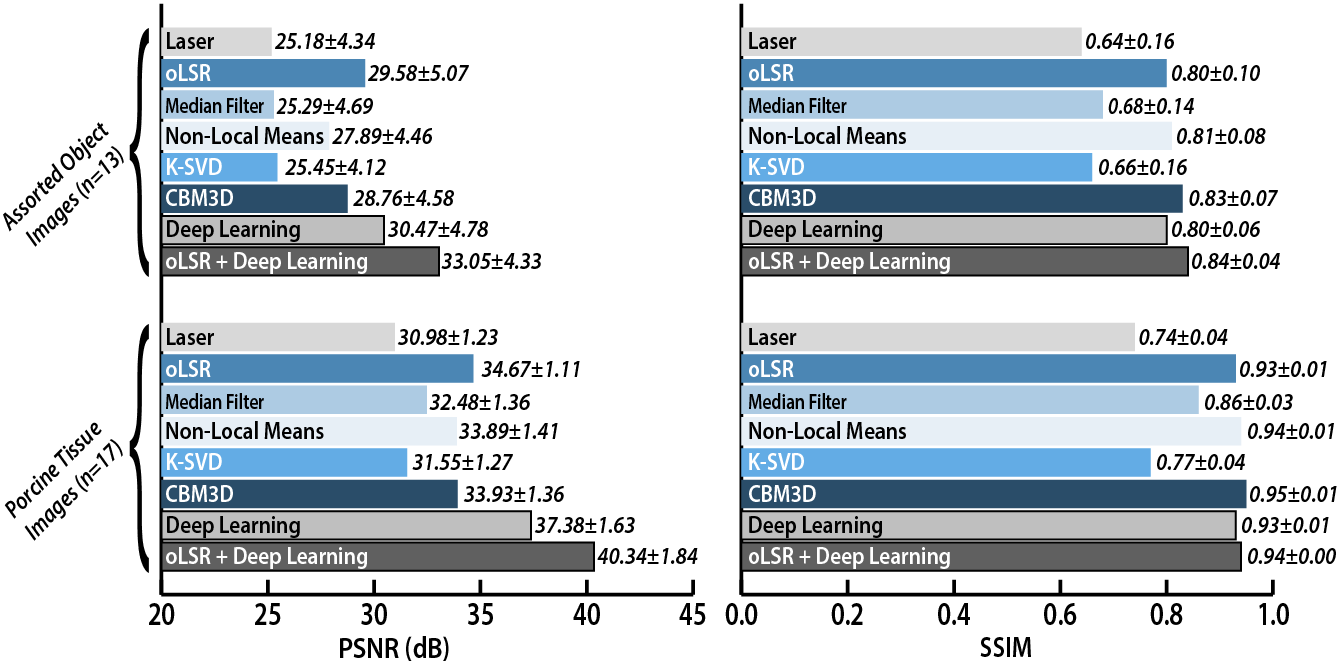}}
\caption{Peak Signal-to-Noise Ratio (PSNR) and Structural Similarity Index (SSIM) results from reserved test images of assorted objects and porcine tissues.}
\label{fig:Table1}
\end{figure}

Validation tests on assorted objects imaged with laser illumination (Fig. \ref{fig:Figure3}a) show that DeepLSR reduces speckle noise by 5.3 dB, compared to a 2.7 dB reduction by non-local means filtering, a 3.6 dB reduction from CBM3D, and a 4.4 dB reduction by oLSR (Fig. \ref{fig:Table1}). We also found that a network trained to transform from input images with optical laser speckle reduction to LED illumination (DeepLSR+oLSR) benefits from greater speckle reduction compared to DeepLSR or oLSR alone. SSIM comparisons between coherent- and incoherent-illuminated images indicate a 25\% improvement in structural similarity for assorted object images and a 26\% improvement for porcine tissue images after applying DeepLSR. While this improvement in recovering object structure is similar to the result of NLM and CBM3D processing, DeepLSR more-effectively reduced speckle noise while maintaining high-frequency image features, demonstrated by the larger PSNR improvement. This result can be seen in NLM- and CBM3D-processed images, which reduce both laser speckle and high-frequency object features, resulting in a blurry appearance of the lines and text in Fig. \ref{fig:Figure3}.

\begin{figure}[t!]
\centering
\makebox[\textwidth][c]{\includegraphics[width=11.5cm]{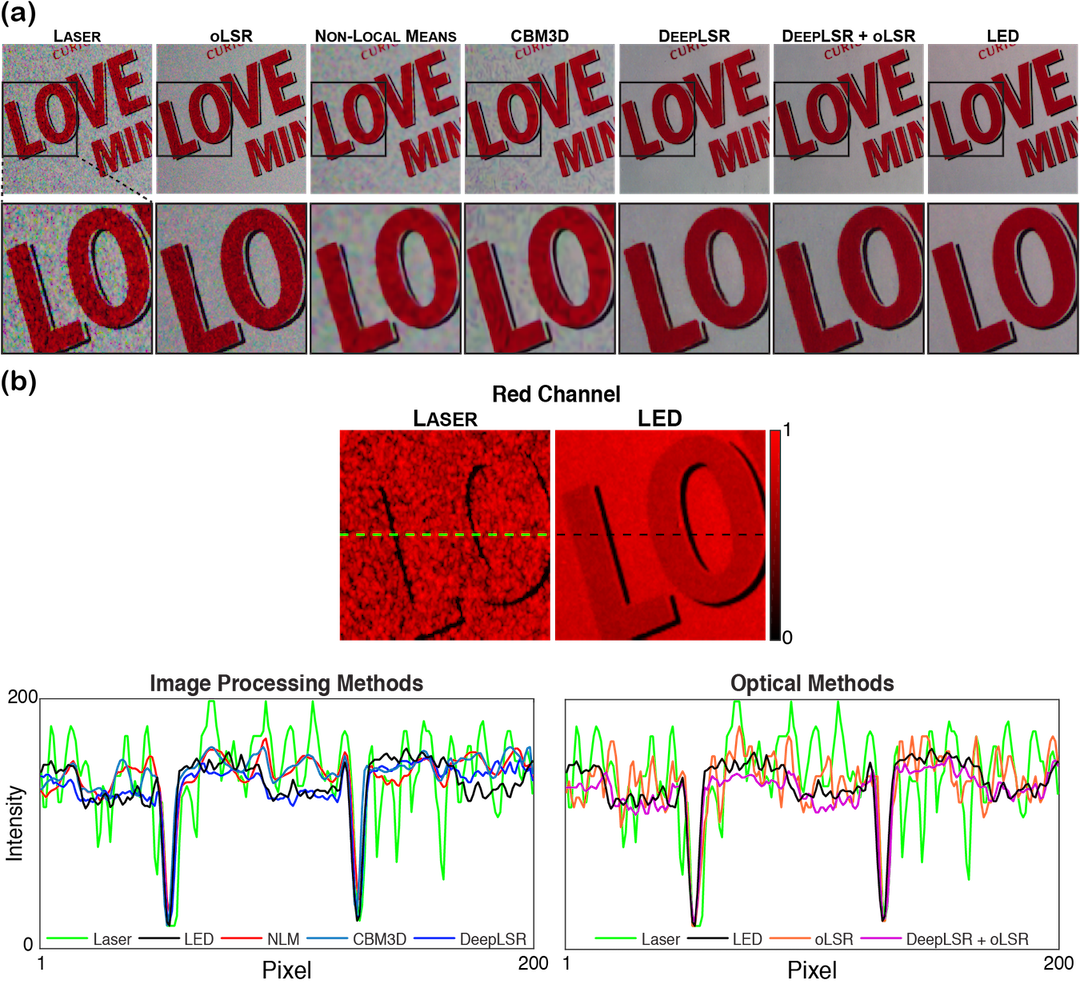}}
\caption{(a) Images of a test object for each speckle reduction technique. (b) The red channels from the color images were studied to assess speckle reduction in the absence of information from other channels. Line profiles from a reserved test image patch are reported, comparing image processing methods (NLM, CBM3D, DeepLSR) and optical methods (oLSR, DeepLSR+oLSR) to the input (Laser) and ground truth (LED) images.}
\label{fig:Figure4}
\end{figure}

\newpage The DeepLSR method has a small effect on resolution measured by a slanted edge test, as demonstrated through the modulation transfer functions of LED illumination compared to laser illumination with DeepLSR. DeepLSR resulted in a reduction in spatial resolution of 17\% when comparing the spatial frequency at which each MTF reached half-modulus (Fig. \ref{fig:Figure3}c). The pixel-level effects of image processing (NLM, CBM3D, DeepLSR) and optical speckle reduction (oLSR, DeepLSR+oLSR) methods are compared in the line profiles shown in Fig. \ref{fig:Figure4}. While non-local means and CBM3D processing suppress high spatial-frequency details of both speckle and original object features, DeepLSR appears to reduce speckle while maintaining edge features of the original object. Lastly, Fig. \ref{fig:Figure4} shows these same trends are observed in object regions that contain primarily red light intensity, indicating that DeepLSR does not rely on information from other color channels for despeckling.

\begin{figure}[t!]
\centering
\makebox[\textwidth][c]{\includegraphics[width=13.335cm]{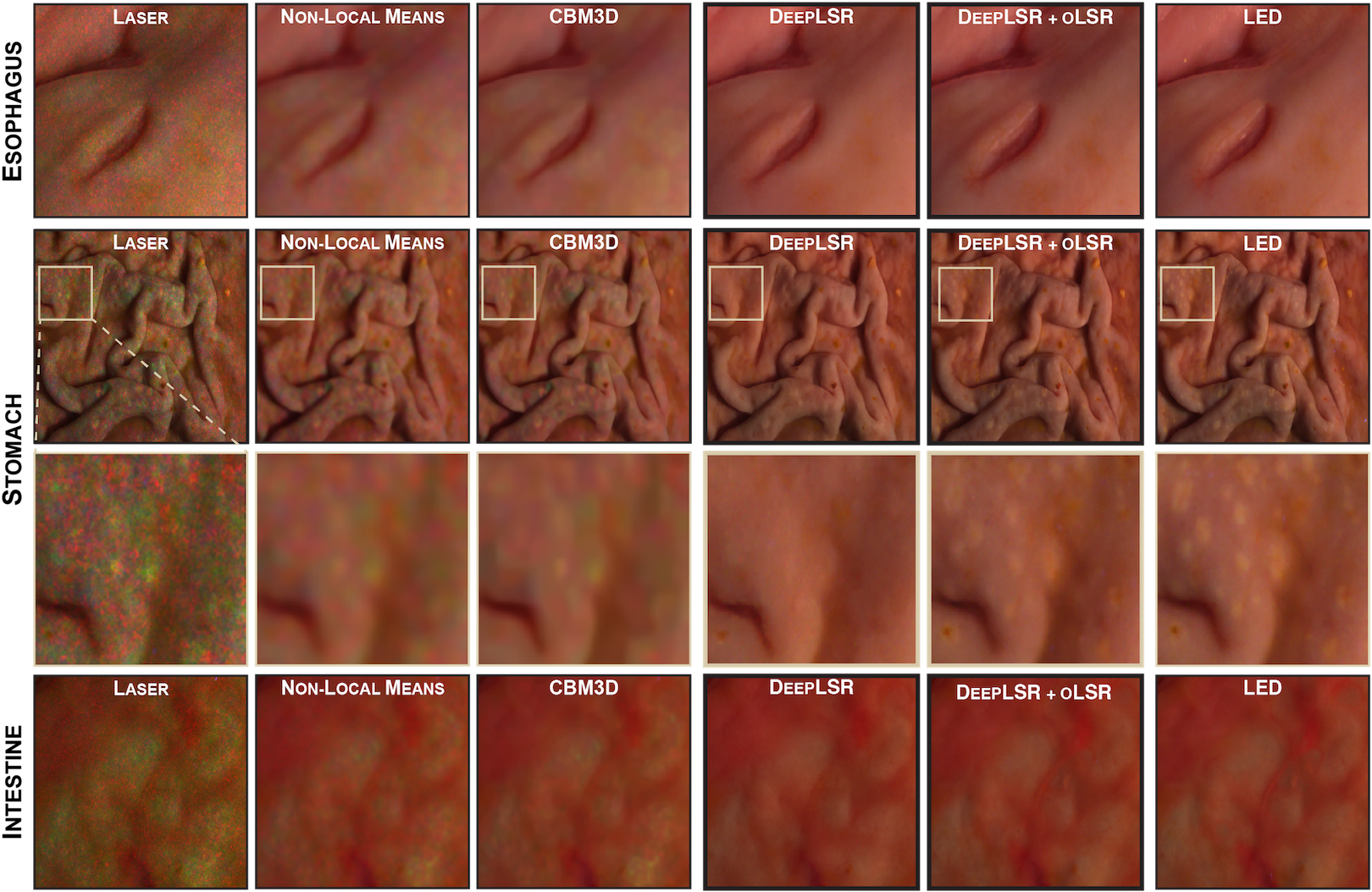}}
\caption{DeepLSR applied to images of laser-illuminated \textit{ex-vivo} porcine gastrointestinal tissues not previously seen by the network.}
\label{fig:Figure5}
\end{figure}

In applications involving tissue imaging, the object of interest is often a turbid medium that naturally blurs speckle artifacts. To assess the applicability of DeepLSR in this scenario, we applied our model to images of gastrointestinal tissue illuminated with laser light. In these tissue validation tests, DeepLSR reduced speckle noise by 6.4 dB, compared to 2.9 dB reduction by non-local means filtering, a 3.0 dB reduction by CBM3D, and a 3.7 dB reduction by oLSR. Fig. \ref{fig:Figure5} shows representative images of test samples with laser illumination, conventional noise reduction methods, DeepLSR, DeepLSR combined with optical speckle reduction, and LED illumination. While CBM3D resulted in only a marginal PSNR improvement in tissue imaging, it did outperform other methods in recovering structural features assessed by SSIM. K-SVD saw minimal improvements in image quality in both imaging expirements. As in the assorted object tests, DeepLSR removed a greater number of speckle artifacts than conventional imaging processing approaches while retaining structural features (Fig. \ref{fig:Table1}). Compared to the assorted object results, the performance metrics from tissue imaging were elevated due to inherent speckle reduction caused by the blurring effect of turbid media.

\section{Discussion}

Compared to many model-based image processing methods, DeepLSR considers large portions of the image\textquotesingle s distribution, allowing it to predict the effective spatial relationships across the image instead of considering only individual pixels. The robustness and generalizability of DeepLSR would improve with increased quantity and variety of training data. Additionally, training data with a larger variety of laser sources, working distances, and imaging lens parameters will improve the generalizability of DeepLSR. To minimize the required quantity of manually captured data, a forward model may be utilized to generate synthetic data with varying speckle size, wavelengths, and small perturbations to prevent the network from learning shape and object texture \cite{PhysMed}. Moreover, future work with unsupervised networks may extend the DeepLSR approach to applications where images with and without speckle noise can be obtained but not registered.

DeepLSR may be applied to virtually any modality in which paired coherent-illuminated and speckle reduced samples can be obtained, such as OCT or ultrasound. Another application that may benefit from DeepLSR is automated quality inspection of products in a factory setting, where remote lighting and bright sources with long lifetimes are required. DeepLSR may be particularly useful in endoscopy applications that require bright illumination or small-diameter endoscopes. Incoherent light sources for endoscopy, such as arc lamps and LEDs, require large-diameter light guides to deliver sufficient optical power through an endoscope. Laser illumination enables the delivery of greater illumination power through fiber optics, and can generate incoherent-like images after DeepLSR or DeepLSR+oLSR is applied. Moreover, in widefield applications that require coherent light, such as laser speckle contrast imaging for mapping flow \cite{Dunn2001}, DeepLSR allows both a computational image and a conventional image to be acquired simultaneously.

\section{Conclusion}

We have presented the application of deep learning to laser speckle reduction by utilizing pairs of images illuminated with coherent and incoherent light for training. DeepLSR removes more speckle artifacts than conventional imaging processing approaches, while retaining structural features. As a data-driven approach, DeepLSR should be trained on images that span the target domain. We make available the nearly 3,000 images collected for training and the models described here so that other researchers may use them to augment their collected training images. We have made the DeepLSR model and source code for widefield laser illumination available here: \url{https://durr.jhu.edu/DeepLSR}, and provide step-by-step instructions for installing and applying this framework to new data sets in the Appendix.

\section{Appendix} \label{sec:Appendix}

\subsection{Instructions for using DeepLSR}

\textit{Setting up cloud computing resources to run DeepLSR} \smallskip\newline
DeepLSR requires performance-capable graphics processing units (GPUs) as training is computationally intensive. We utilized Google Cloud computing to train the models reported in this publication.
\begin{enumerate}
    \item In order to run DeepLSR, setup a Google Cloud instance with the necessary dependencies (\textit{Ubuntu}, \textit{PyTorch}, and \textit{CUDA}) by following the instructions provided here: \url{https://cloud.google.com/deep-learning-vm/docs/pytorch_start_instance}
	\newline
	\colorbox{aliceblue}{%
    \parbox{1\linewidth}{%
	\begin{itemize} 
        \item[$\ast$] \textit{The code released with this publication was tested using Ubuntu 16.04. We recommend utilizing the same operating system to avoid complications.}
        \item[$\ast$] \textit{For best performance, we suggest selecting multiple Nvidia P100 GPUs.}
    \end{itemize}
     	                }
     	                }
    \item With the Google Cloud instance setup, a few dependencies must be installed in order to run the models on a cloud instance. Install \textit{torchvision}, \textit{dominate (v2.3.1+)}, \textit{visdom (v0.1.8.3+)} and \textit{scipy} using the following commands:
	    \begin{itemize}
	    \item For users utilizing \textit{Anaconda}:\newline
		    \code{conda install torchvision -c soumith}\newline
		    \code{conda install -c conda-forge dominate}\newline
		    \code{conda install -c conda-forge visdom}\newline
		    \code{conda install -c anaconda scipy}\newline		    
	    \item For users utilizing \textit{pip}:\newline
 		    \code{pip install torchvision}\newline
		    \code{pip install dominate}\newline
		    \code{pip install visdom}\newline
		    \code{pip install scipy}
	    \end{itemize}
\end{enumerate}

\noindent \textit{Training DeepLSR for laser speckle reduction}
\begin{enumerate}
    \item Begin by cloning the DeepLSR GitHub repository found at (\url{https://durr.jhu.edu/DeepLSR}).
    \item The directory structure for the dataset should be organized as follows:\newline
    \begin{forest}
        for tree={
            font=\ttfamily,
            grow'=0,
            child anchor=west,
            parent anchor=south,
            anchor=west,
            calign=first,
            edge path={
            \noexpand\path [draw, \forestoption{edge}]
            (!u.south west) +(7.5pt,0) |- node[fill,inner sep=1.25pt] {} (.child anchor)\forestoption{edge label};
        },
        before typesetting nodes={
        if n=1
            {insert before={[,phantom]}}
            {}
        },
        fit=band,
        before computing xy={l=15pt},
    }
    [SOMEPATH \# Some arbitrary path
    [Datasets \#Datasets folder
        [XYZ\_Dataset \#Active dataset
    	    [test]
    	    [train]
  	        ]
        ]
    ]
    \end{forest}
    \item To utilize the training data and models referenced in our publication, visit our GitHub repository for download and use instructions (\url{https://durr.jhu.edu/DeepLSR}).
    \item If you intend to train models using your own dataset:
	    \begin{itemize}
    	    \item All test and train data should be in either .jpeg, .jpg or .png formats. .tiff and other raw formats can result in extremely slow training.
	        \item All data must be paired side-by-side i.e. the input and output should be concatenated end-to-end in a single image.
	        \item The size of each individual image should be 2nx2n.
	        \item Once the dataset is setup use the following command for training:\newline
	        \code{python train.py --dataroot $<$datapath$>$--name DeepLSR  --gpu\_ids 0 --display\_id 0 --lambda\_L1 70 --niter 200 --niter\_decay 200 --pool\_size 64 --loadSize $<$image\_size$>$ --fineSize $<$image\_size$>$}
	    \end{itemize}
	\balance
	\colorbox{aliceblue}{%
    \parbox{1\linewidth}{%
	Training Parameters:\newline 
	    \code{--niter} is the number of epochs trained with a constant learning rate 
	    \newline \code{--niter\_decay} is the number of epochs trained with a linearly decaying learning rate.
	    \newline \code{--lr} adjusts the learning rate (default = 0.002).
	    \newline \code{--gpu\_ids} is the number of GPUs used (0 is one GPU, 1 is two GPUs and -1 is no GPU).
	    \newline \code{--lambda\_L1} is the $\ell_1$ regularization parameter used for training. The default is 70 because on our data, the tuned range was [67,74].
	    \newline \code{--load\_Size} is the size of the input/output image.
	    \newline \code{--fine\_size} is the size of the random crop from within the image to introduce jitter.
	    }
	    }
	    
	    \item To view training losses and results, run \code{python -m visdom.server and click the URL http://localhost:8097}. For cloud servers replace \textit{localhost} with your IP.
	    \item To view epoch-wise intermediate training results, visit \textit{./checkpoints/DeepLSR/web/index.html}
\end{enumerate}

\textit{Testing the trained network}
\begin{enumerate}
    \item To test with our pre-trained models, visit our GitHub repository for download instructions.
    \item Once the model has been uploaded to Google Cloud and configured, run the following code: 
    \newline \code{python test.py --dataroot $<$datapath$>$ --name DeepLSR --gpu\_ids 0 --display\_id 0 --loadSize $<$image\_size$>$ --fineSize $<$image\_size$>$}
	    \colorbox{aliceblue}{%
   	    \parbox{1\linewidth}{%
   	    \begin{itemize}
	        \item[$\ast$] To test with our data $<$image\_size$>$ should be set to 1024.
	        \item[$\ast$] The test results will be saved to a html file here: \newline\textit{./results/DeepLSR/test\_latest/index.html}
	    \end{itemize}
	    }
	    }
\end{enumerate}

\section*{Funding}

This work was supported in part with funding from the NIH Trailblazer Award (R21 EB024700).\\

\section*{Disclosures}

TLB (P), FM (P), NJD (P)\\

\end{document}